\documentclass[11pt, a4paper, logo, onecolumn,  copyright,]{googledeepmind}

\usepackage[authoryear, sort&compress, round]{natbib}
\usepackage[T1]{fontenc}
\usepackage[utf8]{inputenc}
\usepackage{todonotes}
\usepackage{algorithm}
\usepackage{algpseudocode}
\usepackage{graphicx}
\usepackage{wrapfig}
\usepackage{subcaption}
\usepackage{geometry}
\usepackage{xcolor}
\usepackage{fancybox}
\usepackage{amssymb} 
\usepackage{xspace}
\usepackage{listings}      

\usepackage[most,skins,theorems]{tcolorbox}

\definecolor{userbg}{RGB}{240, 240, 240}
\definecolor{modelbg}{RGB}{255, 255, 255}
\definecolor{successcolor}{RGB}{25, 150, 25}
\definecolor{failcolor}{RGB}{200, 50, 50}
\usepackage{xcolor}
\usepackage{caption}
\usepackage{listings}

\newtcolorbox[auto counter, number within=section]{definitionbox}[2][]{
colback=blue!0!white,
colframe=blue!30!white,
fonttitle=\bfseries,
title=Definition~\thetcbcounter: #2,#1}

\title{Improving Interactive In-Context Learning from Natural Language Feedback}

\renewcommand{\today}{2025-12-01}
\newcommand{\algoname}{RL\textsuperscript{2}F\xspace}

\author[1,*]{Martin Klissarov}
\author[1,*]{Jonathan Cook}
\author[1,*]{Diego Antognini}
\author[1]{Hao Sun}
\author[1]{Jingling Li}
\author[1]{Natasha Jaques}
\author[1]{Claudiu Musat}
\author[1]{Edward Grefenstette}

\affil[*]{Equal contributions}
\affil[1]{Google DeepMind}

\definecolor{darkpurple}{RGB}{102, 51, 153}
\definecolor{lightpurple}{RGB}{204, 153, 255}
\definecolor{lightblue}{rgb}{0.22,0.45,0.70}
\definecolor{forestgreen}{rgb}{0.24,0.50,0.19}

\definecolor{DeepTeal}{RGB}{20, 110, 110}  
\definecolor{LightTeal}{RGB}{235, 245, 245} 

\newcolumntype{L}[1]{>{\raggedright\arraybackslash}p{#1}}

\usepackage{xcolor}
\newif\ifshowcomments
\showcommentstrue 

\tcbset{
  aibox/.style={
    width=\linewidth,
    top=8pt,
    bottom=4pt,
    colback=LightTeal,
    colframe=DeepTeal,
    colbacktitle=DeepTeal,
    fonttitle=\bfseries,
    enhanced,
    center,
    attach boxed title to top left={yshift=-0.1in,xshift=0.15in},
    boxed title style={boxrule=0pt,colframe=white,},
  }
}
\newtcolorbox{AIbox}[2][]{aibox,title=#2,#1}

\begin{abstract}
Adapting one's thought process based on corrective feedback is an essential ability in human learning, particularly in collaborative settings.
In contrast, the current large language model training paradigm relies heavily on modeling vast, static corpora.
While effective for knowledge acquisition, it overlooks the interactive feedback loops essential for models to adapt dynamically to their context.
In this work, we propose a framework that treats this interactive in-context learning ability not as an emergent property, but as a distinct, trainable skill.
We introduce a scalable method that transforms single-turn verifiable tasks into multi-turn didactic interactions driven by information asymmetry. 
 We first show that current flagship models struggle to integrate corrective feedback on hard reasoning tasks.
 We then demonstrate that models trained with our approach dramatically improve the ability to interactively learn from language feedback. 
 More specifically, the multi-turn performance of a smaller model nearly reaches that of a model an order of magnitude larger.
 We also observe robust out-of-distribution generalization: interactive training on math problems transfers to diverse domains like coding, puzzles and maze navigation.
 Our qualitative analysis suggests that this improvement is due to an enhanced in-context plasticity.
Finally, we show that this paradigm offers a unified path to self-improvement. 
By training the model to predict the teacher's critiques—effectively modeling the feedback environment—we convert this external signal into an internal capability, allowing the model to self-correct even without a teacher.
\end{abstract}

\begin{document}
\maketitle

\section{Introduction}

\begin{wrapfigure}[18]{r}{0.52\textwidth}
    \vspace{-1cm}
    \includegraphics[width=.98\linewidth]{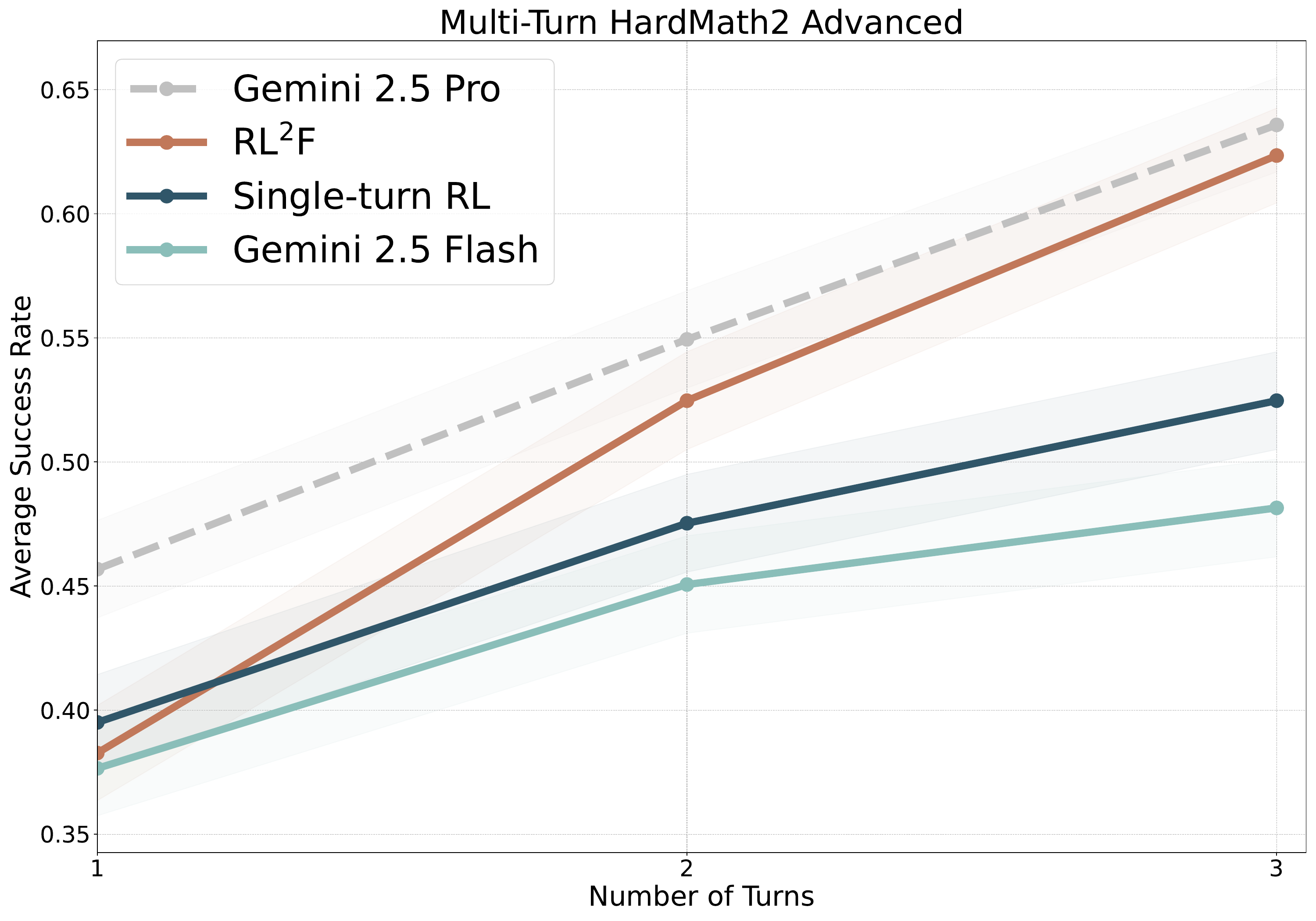}
    \vspace{-0.2cm}
    \caption{Trained through RL on multi-turn didactic interactions,   \textbf{Gemini 2.5 Flash nearly reaches the performance of Gemini 2.5 Pro} on the challenging HardMath2 dataset. Single-turn RL only slightly improves on the baseline's ability to in-context learn from language feedback across turns.}
    \label{fig:v3p1s}
\end{wrapfigure}

\textit{How do we impart capabilities?}
Natural language is central to human progress, allowing us to both distribute information via static sources, like scientific articles, as well as provide real-time feedback to adapt collectively to unpredictable situations. 
While Large Language Models (LLMs) excel at processing large amounts of static knowledge, they struggle to learn from real-time interaction.
Consequently, because models do not readily adapt on the fly, AI researchers must typically intervene when capabilities are missing by manually curating and adding datapoints, or tasks, to the training mixture.
Alternatively, researchers, and users alike, rely on complex prompt engineering to steer models toward desired outcomes.
While effective, these strategies create a bottleneck, placing the burden entirely on researchers to monitor failures and adjust task distributions.
As the tasks users assign to LLMs grow in specificity,
maintaining this level of oversight becomes prohibitively difficult.

\begin{figure}[t!]
    \centering
\includegraphics[width=.98\linewidth]{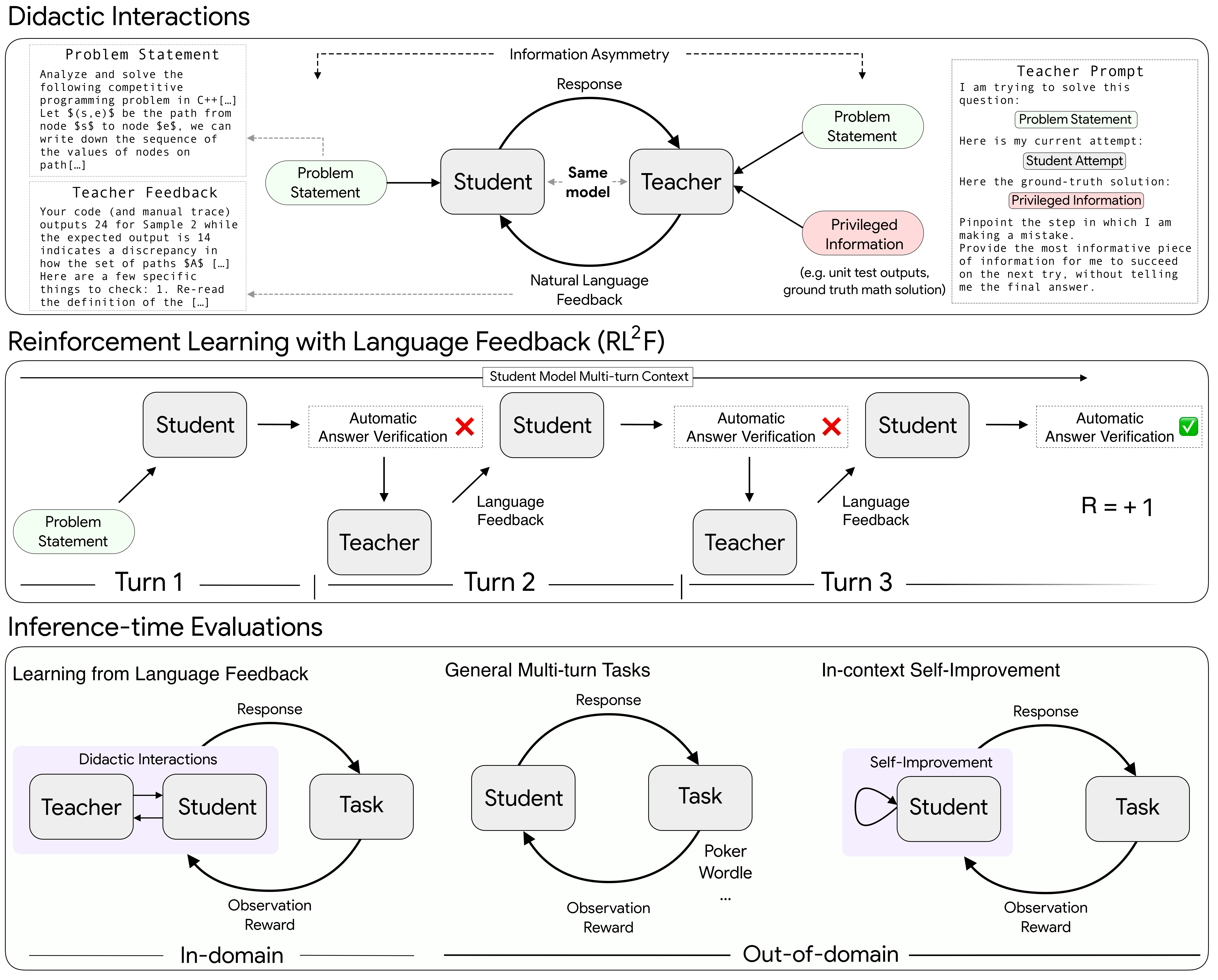}
\caption{
\textbf{(Top) Didactic Interactions via Information Asymmetry:} We transform single-turn problems into multi-turn didactic interactions. A \textit{teacher} model, conditioned on privileged information (e.g., the ground-truth solution), provides natural language feedback to a \textit{student} model without revealing the final answer, guiding it to correct its errors.
\textbf{(Middle) Train-time RL Fine-tuning:} We train the student model to effectively incorporate language feedback using RL. The student iterates through multiple turns; if the answer is correct (checked via automatic verification), a reward ($R=+1$) is granted and the interaction ends. If it is incorrect, the teacher provides feedback. If the interaction reaches the maximum number of turns, $T_{max}$ turns, the reward is zero.
\textbf{(Bottom) Inference-time Evaluations:} We assess the trained model in three settings: (1) \textit{Learning from language feedback} (interacting with an external source of language feedback), (2) \textit{General Multi-turn Tasks} (out-of-domain tasks like logic puzzles or games), and (3) \textit{In-context Self-Improvement}, where the model plays the role of both student and teacher to self-correct.}
    \label{fig:diagram}
\end{figure}
Yet, for many tasks of interest, the knowledge required to resolve these tail-end failures likely already exists, distributed across the model's user base.
In such cases, the bottleneck is not the availability of knowledge, but the model’s capacity to effectively acquire it through interactions.
However, guiding current LLMs via interaction is brittle, often requiring users to engineer complex initial prompts to elicit desired capabilities \citep{White2023A}. In Figure \ref{fig:two_models}, we present this through a simple evocative example between two leading closed-source models, where one of them  misses the obvious hint.
To make it worthwhile for users to teach LLMs, models must adeptly interpret, reason, and adjust their behavior dynamically within the context of a conversation. 
Doing so robustly could lead to exponentially more efficient learning \citep{xu2025provablylearninglanguagefeedback}.
However, this presents a significant challenge: unlike standard Reinforcement Learning (RL) where scalar rewards offer a clear gradient for optimization, language feedback does not readily lend itself to such direct optimization.

In this work, we propose a framework to address this problem by treating interactive in-context learning from natural language feedback as a distinct, trainable capability. 
We introduce a simple and scalable method that reformulates the vast amounts of available single-turn verifiable reasoning problems into multi-turn didactic dialogues. As shown in Figure \ref{fig:diagram}, we simulate interactions between a student and a teacher, where the student must iteratively refine its solution by integrating the teacher's guidance. 
This setup serves as an effective approximation to the real-world corrective feedback between users and models.
A key insight in our work is that \textbf{generating high-quality feedback on verifiable domains does not require a superior model, but rather information asymmetry}.
The teacher can utilize the same base model as the student but is conditioned on privileged information, such as reference solutions or unit test outputs.
By prompting the teacher to provide guidance without revealing the answer (experiments show this happens in less than $1\%$ of cases), we can fine-tune the student model to improve its ability to effectively adapt in-context to natural language feedback. 

\begin{figure}[b!]
    \centering
    
    \footnotesize 
    
    \begin{minipage}[t]{0.48\textwidth}
    \begin{center}
        \textbf{\underline{Model A}}  
    \end{center}
        
        \textbf{Model (Initial Output):} ...
         The statement is a direct summary of the mathematical reasoning presented in the context.
        The context establishes a system of equations using the variables 'x' (for a pencil) and 'y' (for an eraser):
        $3x + y = 1.24$ and $5x + y = 1.82$. It then solves the system by subtracting the first equation from the second to get $2x = 0.58$, which simplifies to $x = 0.29$.
        
        The statement describes this exact process and outcome, merely substituting the variable 'p' for 'x' and 'e' for 'y'. This change in variable names is superficial and does not alter the underlying logic or the result. Therefore, the statement is fully supported and entailed by the context.
        
        \vspace{1em}
        \textbf{User:} \ovalbox{be concise}
        
        \vspace{1em}
        \textbf{Model:} 
        \textcolor{red}{\textit{(Fails to condense content)}}\\
        Got it. I'll keep my responses concise and to the point.
    \end{minipage}
    \hfill 
    \begin{minipage}[t]{0.48\textwidth}
    \begin{center}
        \textbf{\underline{Model B}}  
    \end{center}
        
        \textbf{Model (Initial Output):}...
        Subtracting the first equation from the second eliminates the variable for the eraser, yielding:
        $$ (5p + e) - (3p + e) = 1.82 - 1.24 \Rightarrow 2p = 0.58 \Rightarrow p = 0.29 $$
        This matches the conclusion reached in the context, where $x = 0.29$. Since the structure and outcome of the reasoning are identical, the statement is entailed by the context.
        
        \vspace{1em}
        \textbf{User:} \ovalbox{be concise}
        
        \vspace{1em}
        \textbf{Model:} \textcolor{olive}{\textit{(Successfully summarizes)}}\\ 
        The statement uses different variable names but follows the same logic as the context. Both subtract the equations $3x+y=1.24$ and $5x+y=1.82$ (or $3p+e=1.24$ and $5p+e=1.82$), eliminate the eraser term, and solve $2x=0.58$ (or $2p=0.58$) to find the pencil cost is $0.29$. Thus, the statement is entailed.
    \end{minipage}
    \caption{\textbf{Comparison of the interactive ability between two leading closed-source models.} The user prompts both models to "be concise" after an initial lengthy explanation. \textbf{Model B (Right)} correctly interprets this as a request to summarize the previous explanation. \textbf{Model A (Left)} interprets this as a setting change for future turns and acknowledges the instruction without actually rewriting the content, failing to satisfy the user's obvious intent.}
    \label{fig:two_models}
\end{figure}

 We demonstrate the effectiveness of our approach through a comprehensive series of experiments across diverse domains. 
 We first establish that off-the-shelf flagship models struggle to integrate language feedback during complex reasoning tasks. 
 \textbf{We then study how our didactic training paradigm 
 improves an agent's interactive in-context learning ability} and
 empowers it to actively leverage teacher feedback, turning initial errors into successes through multi-turn dialogue. 
 Remarkably, a smaller model (Gemini 2.5 Flash) fine-tuned via our method nearly matches the performance of a significantly larger model (Gemini 2.5 Pro). 
 Furthermore, we verify the strong generalization of these capabilities: models trained exclusively on verifiable math interactions show marked improvements in distinct domains, such as coding and logic puzzles (e.g., ARC-AGI). 
 This transfer learning even extends to multi-turn tasks unrelated to the didactic training setup, including Poker and Wordle.
 Investigating the mechanism behind these improvements, we find that fine-tuned models exhibit greater in-context plasticity, whereas baselines tend to merely repeat their initial answers.

Finally, we demonstrate that this approach creates \textbf{a unified pathway to self-improvement}. Rather than training solely on the student's turns, we train on the full interactive dialogues, including the teacher's critiques. Considering the teacher as the environment, providing new observations between student turns, this process is analogous to having an auxiliary world modeling objective. Importantly, we train the model to predict the teacher's turns as if they were unconditioned on privileged information, requiring the model to infer the logic of the critique solely from the context of the error. This allows a single model to internalize the whole feedback loop: it learns the student's ability to correct reasoning and the teacher's ability to spot flaws. Consequently, the external guidance provided during training is transformed into an internal capability, allowing the model to self-correct by playing the role of its own teacher.

\section{Learning from Language Feedback}
\label{sec:method}

We now detail our methodology for enhancing the interactive in-context learning capabilities of LLMs. Our design prioritizes verifiable tasks, where performance is measured strictly by the model's ability to achieve the ground-truth solution. This focus allows us to isolate methods that robustly drive improvement across multiple interaction turns, thereby avoiding the evaluation ambiguity and subjectivity inherent in non-verifiable tasks.

Interactive learning via language feedback can be viewed through the lens of social learning \citep{bandura1977social,jaques2019social,ndousse2021emergent}—the process of acquiring behaviors by observing, imitating and interacting with others. Our algorithm builds on one of the simplest yet most powerful social learning settings: the didactic interaction between a teacher and a student.
As illustrated in Figure \ref{fig:diagram}, our mechanism pairs a student model, which attempts to solve a problem, with a teacher model that provides targeted language guidance. 
The student is conditioned on the problem statement; if the automatic verifier deems its answer incorrect, the teacher intervenes with guidance, prompting the student to attempt the problem again.
Unlike common student-teacher frameworks such as distillation \citep{hinton2015distilling,agarwal2023onpolicy}, where the teacher is instantiated as a larger, more capable model, our approach relies solely on information asymmetry. The teacher leverages privileged information (such as ground truth labels) to generate corrective feedback. Consequently, the teacher and student can be instantiated using the exact same model. This cooperative self-play dynamic, optimized via RL, will be shown to yield significant performance improvements after fine-tuning. Our setup is related to \citet{mohtashami2024sociallearningcollaborativelearning}, who introduce a framework of social learning in LLMs where models exchange knowledge via natural language to maintain data privacy. Overall, our teacher-student setup is a particular instantiation of the concepts of language games and boundless socratic learning \citep{schaul2024boundlesssocraticlearninglanguage}.

\subsection{Formal Definition}
\label{formal}
Learning from language feedback can be modeled as a partially observable Markov decision process (POMDP) $\langle \mathcal{S}, \mathcal{A}, \mathcal{O}, T, R\rangle$. The student model, an agent with parameters $\theta_S$, learns a policy $\pi_S$, whereas the teacher model is parametrize with  $\theta_T$ and learns a policy $\pi_T$. In our experiments, the student and the teacher actions $a_t \in \mathcal{A}$ are natural language utterances $u_t$, however, in the general case, this could include tool calls. The observation at time $t$ is modeled as the conversation history $o_t = (u_1^T, u_1^S, \dots, u_t^T, u_t^S)$, where $u_t^T$ and $u_t^S$ are teacher and student utterances at time $t$.
A state at time $t$ is then defined as the tuple consisting of the conversation history $o_t$ and the task's privileged information $k$, written as  $s_t = (k_t, o_t) = (k_t, u_1^T, u_1^S, \dots, u_t^T, u_t^S) \in \mathcal{S}$. This privileged information can take the form of a ground-truth solution or output of private unit tests.

 At each turn $t$, the teacher observes the conversation history and the task privileged information $k$, and  takes an action $u_{t+1}^T \sim \pi_T(\cdot| s_t,k)$.
After the teacher's utterance, the student responds with $u_{t+1}^S \sim \pi_S(\cdot | u_{t+1}^T, o_t)$. In our implementation, the teacher and the student policies define the transition function $T$. In the general case, these policies could be acting on an environment (such as computer use) with a separate transition function.
This interaction loop continues until the end of an episode, defined through a maximum number of turns $T$, or when the student outputs the correct solution. 
The reward function then measures the student's output through a sparse function $R(s_t, a_t)$ focusing on conversation-level outcomes like final answer correctness. We use verifiable domains like math and code, where objective performance measures exist.

Our algorithm leverages the model's in-context learning ability to integrate natural language feedback during conversations. After each such trajectory, weight optimization is done through gradient updates. In other words, we are improving the model's ability of \textit{learning how to learn from interactions}, i.e. its meta-learning ability \citep{Schmidhuber1987EvolutionaryPI,Thrun1998LearningTL}.
The standard approach in meta learning, recently generalized by recent works \citet{pmlr-v80-franceschi18a,Grefenstette2019Generalized} through the lens of bi-level optimization, typically requires gradient updates in the inner loop, leading to meta-gradients.
 We avoid this by relying on, and improving, the existing in-context learning abilities of LLMs. 
 
 This approach is similar to black box meta-learning (e.g., RL$^2$) \citep{wang2017learningreinforcementlearn,duan2016rl2fastreinforcementlearning} and first-order meta-learning \citep{nichol2018first}. In particular, since during training language feedback is only observed following turns that do not achieve the verifiable reward, the presence of language feedback can be interpreted as an augmented reward observation. The LLM can thus be viewed as implementing an in-context RL algorithm with weights that are in turn optimized via RL. We name our method \textbf{Reinforcement Learning with Language Feedback (RL$^2$F)} to indicate the use of these two learning signals and to highlight the connection with RL$^2$.


\section{Experiments}
We perform an extensive study designed to measure the efficacy and  generalizability of our proposed method for improving the interactive in-context learning abilities of LLMs. We structure our investigation into three distinct areas:

\begin{enumerate}
    \item We first analyze the interactive in-context learning abilities of state-of-the-art models, revealing \textbf{significant headroom in the frontier models' multi-turn reasoning abilities} across interactions on hard reasoning tasks.
    \item We conduct fine-tuning experiments on a non-thinking model, Gemma 3 12b, as well as the thinking flagship Gemini 2.5 Flash. Experiments focus on measuring the improvements in terms of multi-turn in-context adaptation, where \textbf{\algoname}  \textbf{leads to significant in-distribution gains and strong out-of-distribution generalization}. We study the reasons behind such improvements through qualitative analysis, which points to an \textbf{increased in-context plasticity}.
    \item We demonstrate a \textbf{pathway to self-improvement} by training on multi-turn attempts to internalize the feedback loop. By treating the teacher's turns as an auxiliary world-modeling objective, the model can internalize the feedback loop, thereby developing the capability to self-correct.
\end{enumerate}

\subsection{In-depth Evaluation}
\label{sec:eval}
We begin our experiments by evaluating where frontier models stand in terms of their ability to in-context learn from language feedback. We do not perform weight updates in this section, but simply ask: can current models already efficiently incorporate feedback in-context? 
We experiment across four different hard reasoning tasks: HardMath2 \citep{roggeveen2025hardmath2}, ARC-AGI \citep{chollet2024arc}, Codeforces \citep{codeforces} and BIG-Bench Extra Hard \citep{kazemi-etal-2025-big}. We evaluate four separate models, three from the Gemini 2.5 family of models  \citep{gemini25_2025} (Gemini 2.5 Pro, Gemini 2.5 Flash and Gemini 2.5 Flash Lite) and additionally compare to the performance of GPT-5 \citep{openai2025gpt5}.

\begin{figure}[ht!]
    \centering
    \begin{subfigure}[b]{0.48\textwidth}
        \centering
        \includegraphics[width=\linewidth]{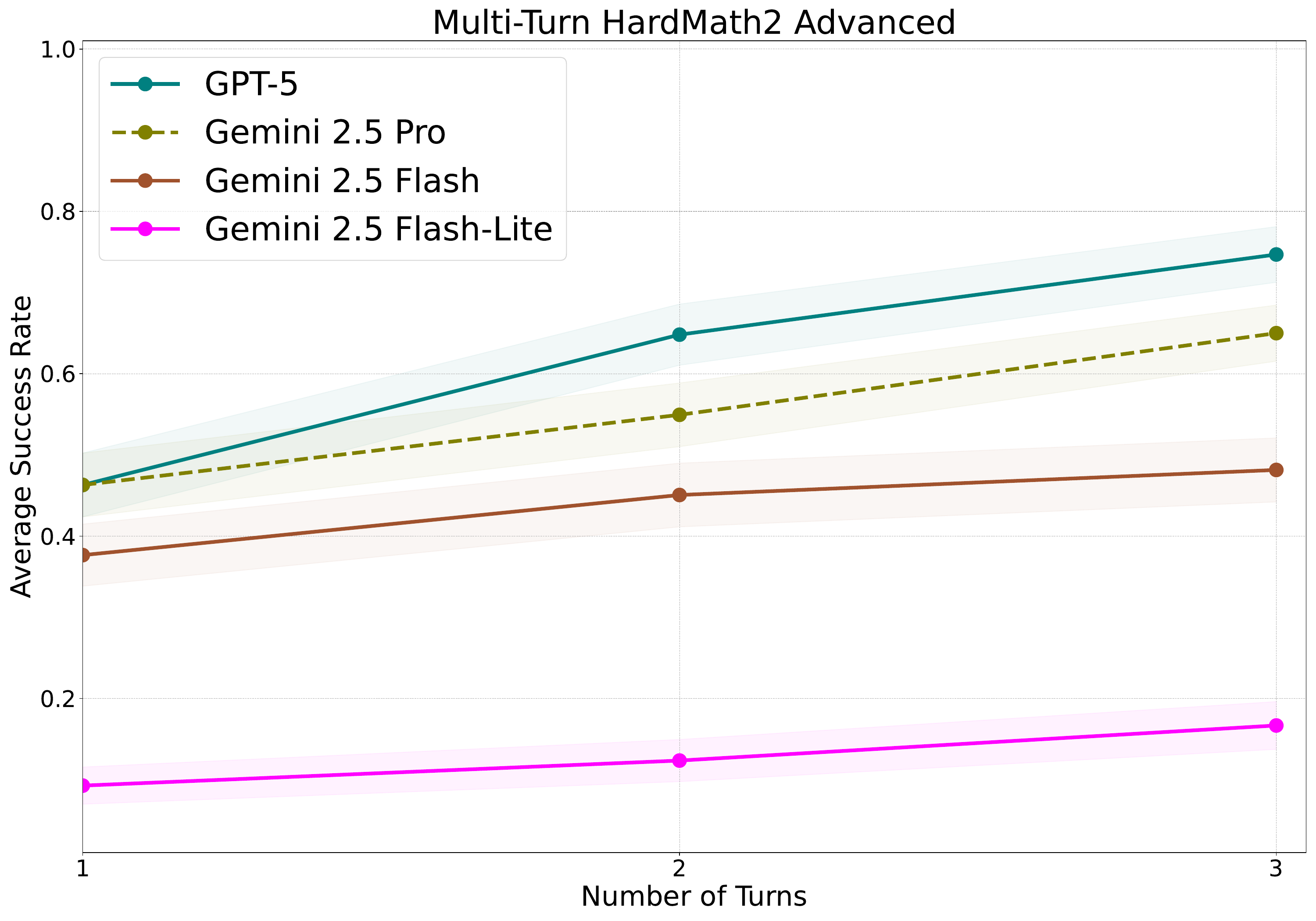}
        \label{fig:sub1}
    \end{subfigure}
    \hfill 
    \begin{subfigure}[b]{0.48\textwidth}
        \centering
        \includegraphics[width=\linewidth]{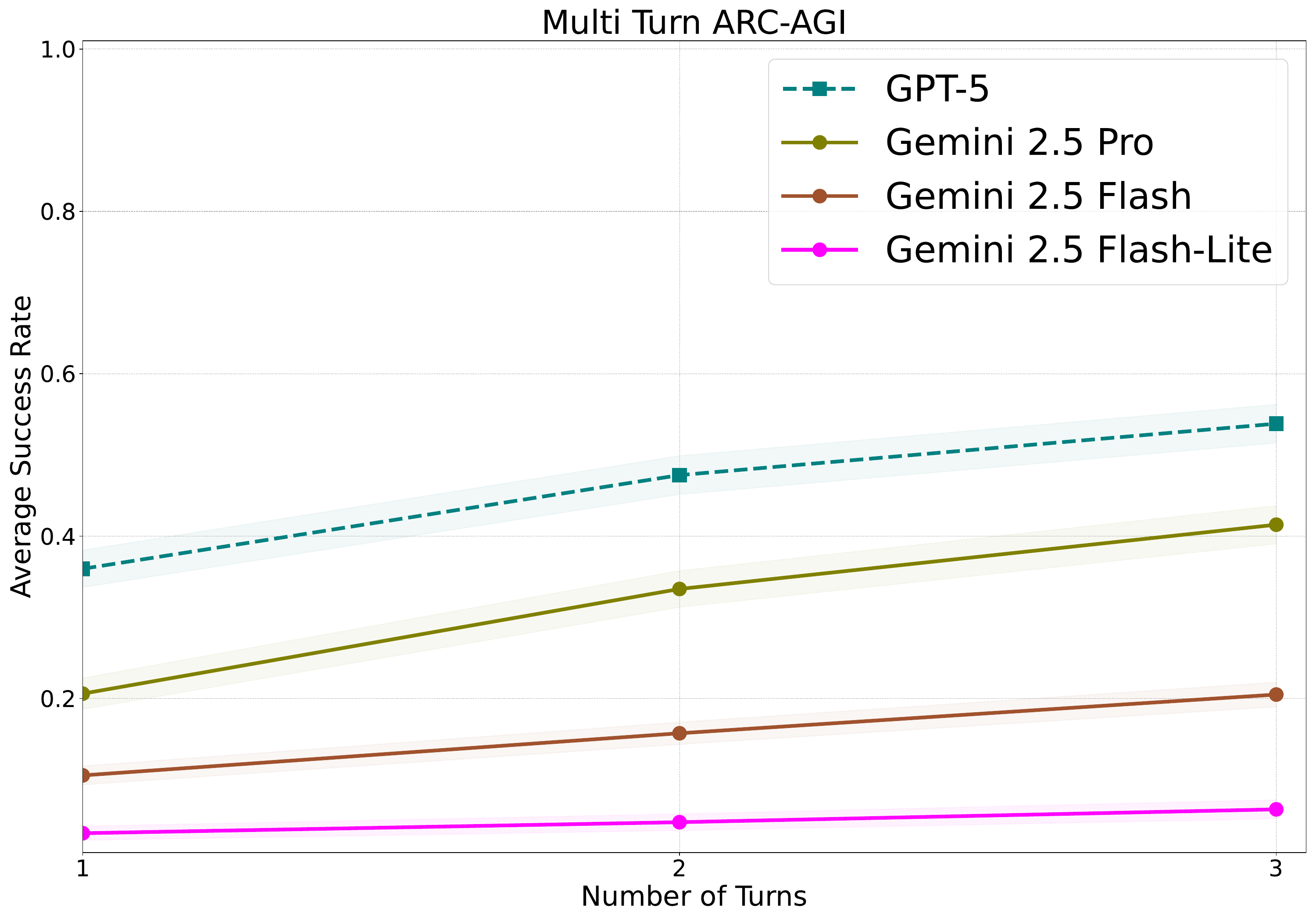}
        \label{fig:sub2}
    \end{subfigure}
    \hfill 
    \begin{subfigure}[b]{0.48\textwidth}
        \centering
        \includegraphics[width=\linewidth]{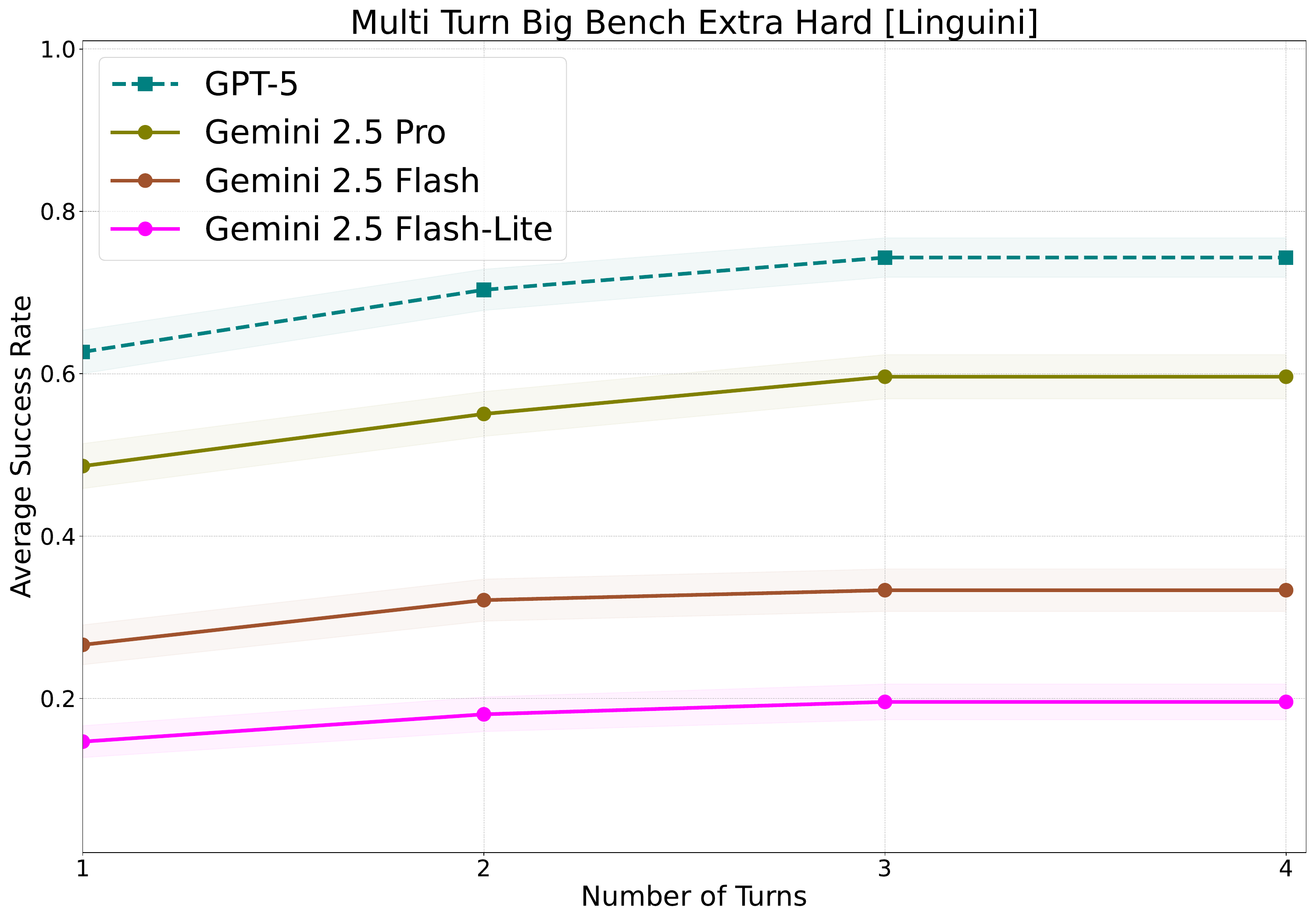}
        \label{fig:sub3}
    \end{subfigure}
    \hfill 
    \begin{subfigure}[b]{0.48\textwidth}
        \centering
        \includegraphics[width=\linewidth]{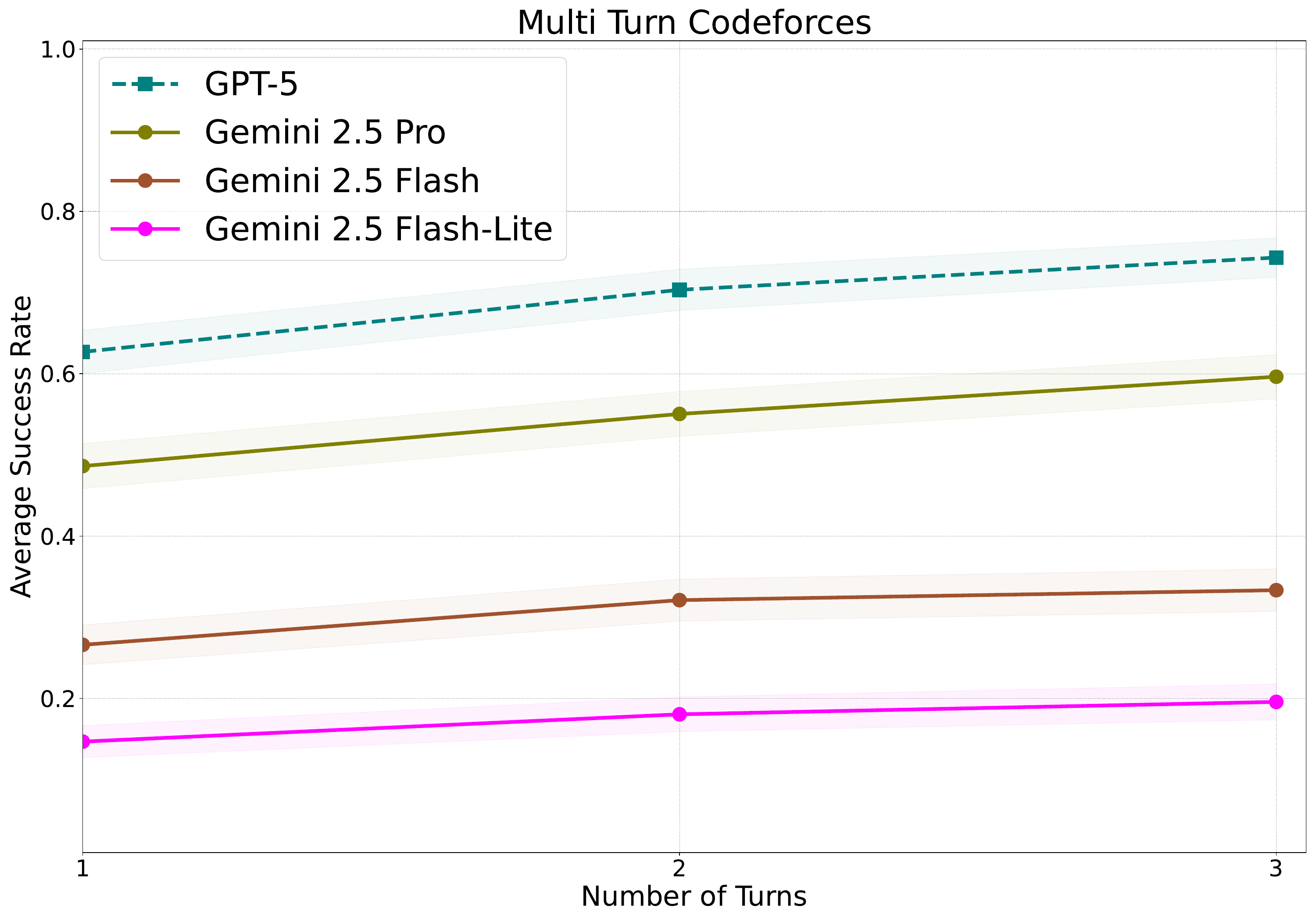}
        \label{fig:sub4}
    \end{subfigure}
    \caption{Cumulative accuracy across four hard reasoning domains: HardMath2, ARC-AGI, BBEH, and Codeforces. \textbf{The plots illustrate the limited ability of the Gemini 2.5 models and GPT-5 to incorporate language feedback over multiple turns.} We observe this ability scales with model size, with GPT-5 generally exhibiting stronger performance gains compared to the Gemini 2.5 Pro model.}
    \label{fig:eval}
\end{figure}

The results in Figure \ref{fig:eval}, we report the cumulative accuracy which is defined as the percentage of problems that were correctly solved for a certain number of turns, shown on the $x$-axis. We observe an ordering between model sizes, where the larger Gemini models tend to perform better than smaller models. Interestingly, we observe that GPT-5 consistently performs better than Gemini 2.5 Pro. This is apparent even in cases where the initial single-turn accuracy is almost the same (on HardMath2). Overall, all models show, that despite multiple turns of feedback from a teacher, the accuracy of the student leaves a lot of room for improvement.

\textbf{Does the teacher leak the ground truth solution?} An important question that we need to address in the current evaluation setup relates to whether we can trust that the student legitimately solves the problem, or has the solution been revealed by the teacher. We investigate this question by having a verification pipeline of two detection mechanisms: (1) a string matching function applied on each teacher output and (2) an LLM judge that, conditioned on the question, ground truth solution and teacher output, evaluates whether the solution has been revealed. We report that in an average of $\textbf{0.3\%}$ of cases our verification pipeline signals that the teacher has revealed the solution, indicating a strong enough instruction-following ability of the Gemini and GPT models. The noise that might come from such false positives is far below the gap in performance between the models that we have evaluated.

\subsection{Didactic Interactions}

Having characterized the gap in interactive in-context learning capabilities, we now present the results on improving this capability through the proposed approach outlined in Section \ref{sec:method}.
We compare this approach to two additional baselines: (1) a standard supervised fine-tuning (SFT) baseline trained on problem and solution pairs and (2)  single-turn RL, the standard optimization method on verifiable tasks, also known as RL from verifiable rewards (RLVR) or RL from machine feedback  (RLMF).

\begin{figure}[ht!]
    \centering
    \begin{subfigure}[b]{0.48\textwidth}
        \centering
        \includegraphics[width=\linewidth]{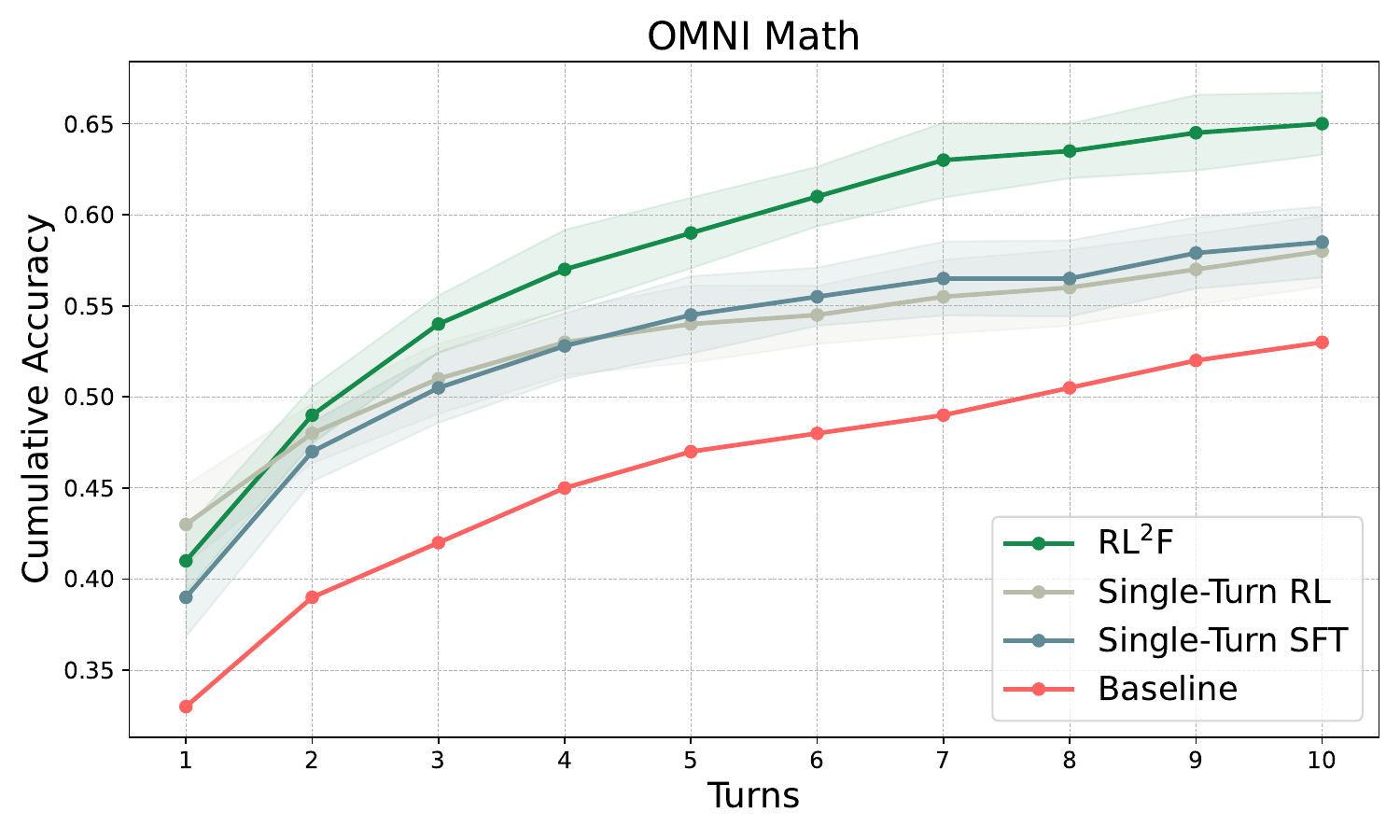}
        \caption{}
        \label{fig:social_learning}
    \end{subfigure}
    \hfill 
    \begin{subfigure}[b]{0.48\textwidth}
        \centering
        \includegraphics[width=\linewidth]{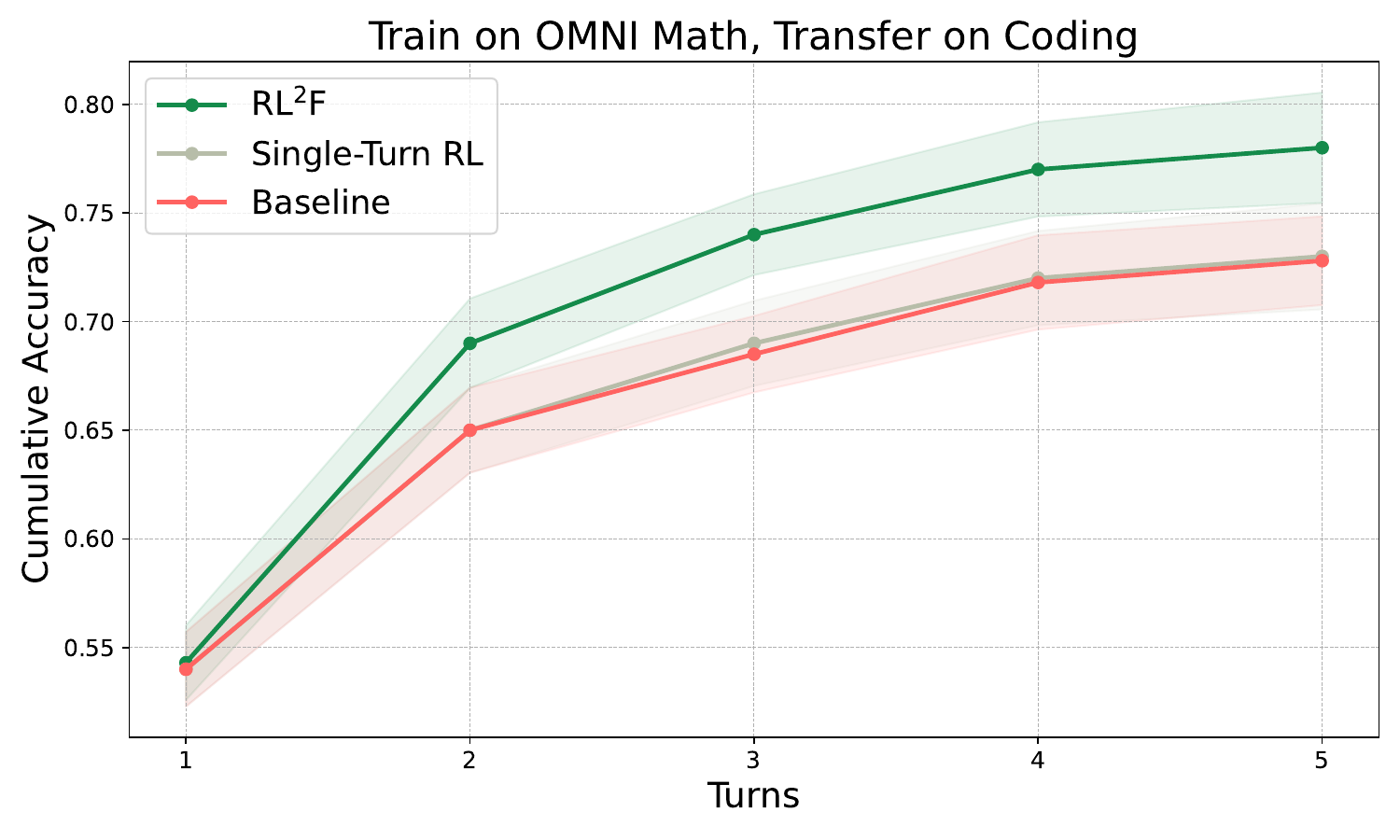}
        \caption{}
        \label{fig:transfer}
    \end{subfigure}
    \caption{\textbf{\algoname on multi-turn didactic interactions is key to improve the interactive in-context learning abilities of LLMs}. The improved performance additionally transfers significantly better to out-of-distribution tasks than single-turn RL.}
    \label{fig:omni_learning}
\end{figure}

\subsubsection{Non-Thinking Models} In our first set of experiments, we fine-tune a non-thinking model, Gemma 3 12b, on the training split of Omni  MATH \citep{Gao2025Omni}. We then evaluate the ability of the model to integrate teacher feedback on the test set and compare performance against a set of baselines.
For Omni MATH, we formed a test split by randomly selecting 500 questions from the larger dataset.
As illustrated in Figure \ref{fig:social_learning}, \algoname on didactic interactions proves to be the most effective strategy for enhancing interactive in-context learning. Crucially, while standard RL and SFT baselines match the single-turn problem-solving improvements, only \algoname continuously widens the performance gap relative to the baselines as the number of interaction turns increases. These findings highlight an effective method for improving reasoning through conversational feedback, offering a natural alternative to the impractical task of curating massive datasets of expert behavior.

Next we verify how training for reasoning from multi-turn language feedback generalizes to out-of-distribution domains. We train the Gemma 3 12b model on Omni MATH and evaluate on LiveCodeBench \citep{Jain2025LiveCodeBench}. 
Here, the teacher’s private knowledge consists of unit test cases. By providing feedback based on test execution results, the system mimics a standard software engineering workflow.
Results in \ref{fig:transfer} show that the single-turn RL baseline barely improves upon the baseline, whereas the multi-turn RL optimization done on teacher-student interactions provides improvements for later turns. This is a first sign indicating that \textbf{learning to learn from language feedback, \algoname, could be more generalizable than standard RLVR}. In the following section, results on thinking models will show a significantly wider spectrum of generalization.

\subsubsection{Thinking Models} 

The previous experiments in this study focused on a relatively small, non-thinking models. It is natural to question whether learning from multi-turn interactions remains necessary for thinking models, given that mechanisms such as language critiques and revisions are already inherent in thinking traces \citep{deepseekai2025deepseekr1incentivizingreasoningcapability}. To address this, we conduct experiments on the Gemini 2.5 Flash model, fine-tuning it via RL on didactic multi-turn interactions. We utilize a private set of mathematics questions whose difficulty exceeds that of the Omni-MATH dataset. We then evaluate performance on a separate and challenging dataset, HardMath2 \citep{roggeveen2025hardmath2}. The results in Figure \ref{fig:v3p1s} demonstrate that \algoname yields substantial gains in improving a model's ability to in-context learn from language feedback. Notably, \textbf{the performance of the fine-tuned Flash model nearly matches that of the Gemini 2.5 Pro model}. This is a significant finding, as mathematics is a domain where Gemini 2.5 Pro typically excels. We show that by eliciting a thinking model's multi-turn capabilities through a training procedure with modest computational cost, we can bridge the performance gap to a model a full tier above.

\begin{table}[htbp]
    \centering
    \small
    \caption{Gemini 2.5 Flash, trained solely on math, is evaluated on three out-of-domain multi-turn benchmarks: ARC AGI, Codeforces, and Linguini. \textbf{This experiment evaluates the ability of the trained models to in-context learn from language feedback by interacting with a teacher model at inference time}. \algoname provides the strongest gains with an average of $7\%$ increase in performance across tasks.}
    \label{tab:in_domain_generalization}
    \begin{tabular}{l c c c}
        \toprule
        \textbf{Tasks} & \textbf{\algoname} & \textbf{Single turn RL} & \textbf{Gemini Flash 2.5} \\
        \midrule
         Multi-turn ARC AGI & \textbf{23.56} & 20.47 & 20.47 \\
        Multi-turn Codeforces & \textbf{37.03} & 32.77 & 33.33 \\
        Multi-turn Linguini & \textbf{56} & 42.35 & 42 \\
        \midrule
        \textbf{Average Performance} & \textbf{38.86} & 31.86 & 31.93 \\
        \bottomrule
    \end{tabular}
    \label{table:ts_generalization}
\end{table}

We also investigate whether the interactive reasoning capabilities learned through \algoname generalize beyond the mathematical domain on which the model was trained.
To test this, we evaluate the math-trained Gemini 2.5 Flash model on three separate benchmarks: ARC AGI  for abstract reasoning, Codeforces for competitive programming, and Linguini (from BIG-Bench Extra Hard) for linguistic logic.
The results, summarized in Table \ref{table:ts_generalization}, reveal a stark contrast between optimization strategies. The single-turn RL baseline fails to generalize, yielding performance outcomes  that are effectively indistinguishable from the baseline Gemini 2.5 Flash. In contrast, \algoname demonstrates strong cross-domain transfer, boosting by nearly $7\%$ on average performance. 

\begin{figure}[ht!]
    \centering
    
    \footnotesize 
    
    \begin{minipage}[t]{0.48\textwidth}
    \begin{center}
        \textbf{\underline{\algoname fine-tuned Gemini 2.5 Flash}}
    \end{center}
        
        \textbf{Question:} Solve the following nonlinear partial differential equation $$ \partial_t u + u^2 \partial_x u = \partial_{xx}u - (x-t)^{-2} $$ [...] and determine its leading-order behavior as $x-t \to \infty$.
        
        \vspace{0.5em}
        \textbf{Model:} \textcolor{red}{\textbf{(Wrong Answer)}}\\
The travelling wave solution is of the form $u(x,t) = U(x-t)$.
We need to compute the partial derivatives in terms of derivatives of $U$ with respect to $\xi = x-t$...\\
Substituting these into the PDE, we get
$-U' + U^2 U' = U'' - (x-t)^{-2}$
The term $(x-t)^{-2}$ becomes $\xi^{-2}$ since...
\begin{lstlisting}
-1/(x-t)
\end{lstlisting}

        \vspace{0.5em}
        \textbf{Teacher:} Your mistake lies in assuming that the constant of integration immediately implied that the solution should tend to zero...
        
        \vspace{0.5em}
        \textbf{Model:} \textcolor{red}{\textbf{(Wrong Answer)}} The hint says solutions can tend to non-zero constants.
So we should investigate solutions such that $U(\xi) \to c \ne 0$ as $\xi \to \infty$...
The original PDE has a nonlinear term $u^2u_x$.
I should look for stationary solutions to something like...
\begin{lstlisting}
(sqrt(5)-1)/2
\end{lstlisting}

        \vspace{0.5em}
        \textbf{Teacher:} You made a mistake when you assumed your solutions had an asymptotic expansion of the form $u(x,t) \sim a + \frac{b}{x-t}$. While for some solutions this might be true, for other solutions this may not be the case...
        
        \vspace{0.5em}
        \textbf{Model:} \textcolor{olive}{\textit{(Successfully integrates feedback)}} The hint asked me to reconsider the possibility of $u(x,t) \to 1$ as $x-t \to \infty$.
The equation is $U' - \frac{1}{3}U^3 + U = -\frac{1}{\xi} + \frac{2}{3}$.
Let's try to find the behavior of $U(\xi)$ near 1...
\begin{lstlisting}
1 + 1/((x-t)**(1/2))
\end{lstlisting}
    \end{minipage}
    \hfill 
    \begin{minipage}[t]{0.48\textwidth}
    \begin{center}
        \textbf{\underline{Baseline Gemini 2.5 Flash}}        
    \end{center}
        
        \textbf{Question:} Solve the following nonlinear partial differential equation $$ \partial_t u + u^2 \partial_x u = \partial_{xx}u - (x-t)^{-2} $$ [...] and determine its leading-order behavior as $x-t \to \infty$.
        
        \vspace{0.5em}
        \textbf{Model:} \textcolor{red}{\textbf{(Wrong Answer)}}\\
        Substituting these expressions into the PDE, we get:
$$ -U' + U^2 U' = U'' - z^{-2} $$
This is a nonlinear ordinary differential equation for $U(z)$. We want to solve this ODE [...]
\begin{lstlisting}
-(x - t)**(-1)
\end{lstlisting}

        \vspace{0.5em}
        \textbf{Teacher:} The step where you assumed $U(z)$ to be small for large $z$ is likely where you went wrong. ... Consider the behavior of solutions to $U'' + (1-U^2)U' = 0$ for large $z$.
        
        \vspace{0.5em}
        \textbf{Model:} \textcolor{red}{\textbf{(Exact Repetition)}} ...I think I should stick with my answer, the argument against $U(z)$ being small was based on considering the homogeneous equation. But the forcing term $z^{-2}$ has significantly affected the asymptotic behavior. I don't see any error in assuming $U(z)$ is small
...
       
\begin{lstlisting}
-(x - t)**(-1)
\end{lstlisting}

        \vspace{0.5em}
        \textbf{Teacher:} 
...When you calculated $U''+U'(1-U^2)$ for $U=-1/z$, you got $z^{-2}-2z^{-3}-z^{-4}$.
You should try to modify the leading order solution to cancel the term $-2z^{-3}$.
The term $-2z^{-3}$ appears from $U''$. Try to find a correction term in $U$ such that it creates a term of $2z^{-3}$ when taking the second derivative to cancel $-2z^{-3}$.
        
        \vspace{0.5em}
        \textbf{Model:} \textcolor{red}{\textbf{(Exact Repetition)}}
[no thinking tokens]
        
\begin{lstlisting}
-(x - t)**(-1)
\end{lstlisting}
    \end{minipage}
    \caption{Conversation comparison between the baseline, the Multi-Turn RL fine-tuned model, and the 2.5 Flash model. \textbf{The \algoname fine-tuned model exhibits greater in-context plasticity, exhibiting greater responsiveness to the teacher feedback and attempting to integrate it}. In comparison, the baseline simply disregards the feedback, constantly repeating the same initial solution. In the last turn, the base model simply gives up on thinking and directly outputs the same answer.}    \label{fig:conversation_comparison}
\end{figure}
\textbf{What could be the cause of such increased performance?} We investigate this question in Figure \ref{fig:conversation_comparison}, which contrasts the behavioral dynamics of the baseline model against the \algoname fine-tuned model on a challenging problem from HardMath2. The analysis reveals a critical difference: despite receiving precise mathematical corrections, the baseline refuses to update its stance, repeating the identical incorrect code and eventually ceasing to utilize thinking tokens entirely. In sharp contrast, the \algoname model demonstrates the ability to critically evaluate external feedback. It utilizes its thinking traces not merely to justify its prior output, but to reason about the teacher's hints, successfully integrating them to derive a new, correct asymptotic expansion. We refer to the baseline inability to integrate feedback and improve its response as a possible \textbf{lack of in-context plasticity}. Plasticity is well-studied phenomenon in RL \citep{dohare,nikishin2022primacy,lyle2023understanding} and is defined as the ability of a neural network to change its predictions in response to new information. However, it is not usually studied beyond the in-weight updates, in particular for the case of in-context learning in LLMs. We believe this work presents a first clear sign of this phenomenon, requiring deeper investigations. In the Appendix \ref{sec:appendix}, we report additional examples, showing how the better integration of feedback naturally leads to task decomposition and improved final performance.

\subsection{Beyond Didactic Interactions}

So far we have seen generalization by modifying the specific domain underpinning the didactic interactions between the teacher and the student. In this section, we investigate whether interactive reasoning capabilities, acquired solely through dialogue on mathematics, transfer to general-purpose multi-turn tasks. We assess the fine-tuned Gemini 2.5 Flash model across a diverse suite of environments, ranging from game playing (Poker and Wordle) to spatial reasoning (Maze Navigation and Only Connect Wall) and user-centric dialogue (Eliciting User Preferences and Movie Recommendations).

\begin{table}[htbp]
    \centering
    \small
    \caption{ We evaluate on 10 diverse out-of-distribution general multi-turn environments, \textit{ranging from game playing to spatial reasoning}. In these tasks, the trained models do not interact with a teacher: they simply act within the agentic environment defined by the tasks. \textbf{Training Gemini 2.5 Flash through \algoname on math shows strong out-of-distribution generalization on multi-turn agentic tasks}, outperforming baselines on 7 out of 10 tasks and achieving 5\% average performance boost. }
    \label{tab:task_comparison}
    \begin{tabular}{l c c c}
        \toprule
        \textbf{Multi-turn Tasks} & \textbf{\algoname} & \textbf{Single turn RL} & \textbf{Gemini 2.5 Flash} \\
        \midrule
        Aider Polyglot Code Edit  & 18.67 & 12.15 & \textbf{21.33}\\
        Circuit Decoding  & 26.42 & \textbf{31.85} & 17.92\\
        Eliciting User Preferences  & \textbf{32.50} & 27.40 & 29.32\\
        Maze Navigation  & \textbf{87.50} & 78.35 & 75.00\\
        Movie Recommendations  & \textbf{69.61} & 68.22 & 69.34\\
        Only Connect Wall  & \textbf{72.00} & 44.75 & 53.00\\
        Poker  & \textbf{38.71} & 36.95 & 36.82\\
        Twenty Questions  & \textbf{95.50} & 91.60 & 91.75\\
        Word Chaining  & 35.50 & 34.68 & \textbf{36.50}\\
        Wordle  & \textbf{59.03} & 57.42 & 56.72\\
        \midrule
        \textbf{Average Performance} & \textbf{51.65} & 46.54 & 46.92\\
        \bottomrule
    \end{tabular}
    \label{table:generalization}
\end{table}
The results, detailed in Table \ref{tab:task_comparison}, demonstrate remarkable cross-domain transfer. Despite being trained exclusively on a limited dataset of math interactions, the \algoname model outperforms both the single-turn RL baseline and the original base model in the majority of domains (7 out of 10), with an average performance boost of nearly 5\%. Notably, the model achieves these gains in unrelated tasks such as Maze Navigation (+12.5\%) and Only Connect Wall (+19\%), suggesting that the ability of learning to learn from natural language feedback, cultivated through multi-agent interactions, is a fundamental cognitive capability that generalizes far beyond the source domain. Given the negligible cost of synthesizing these interactions, this approach offers a highly data-efficient pathway for scaling post-training optimization in thinking models by re-purposing all available single-turn verifiable tasks to train the model's interactive in-context learning ability.

\section{Towards Self-Improvement}

\begin{figure}[ht!]
    \centering
    \begin{subfigure}[b]{0.57\textwidth}
        \centering
        \includegraphics[width=\linewidth]{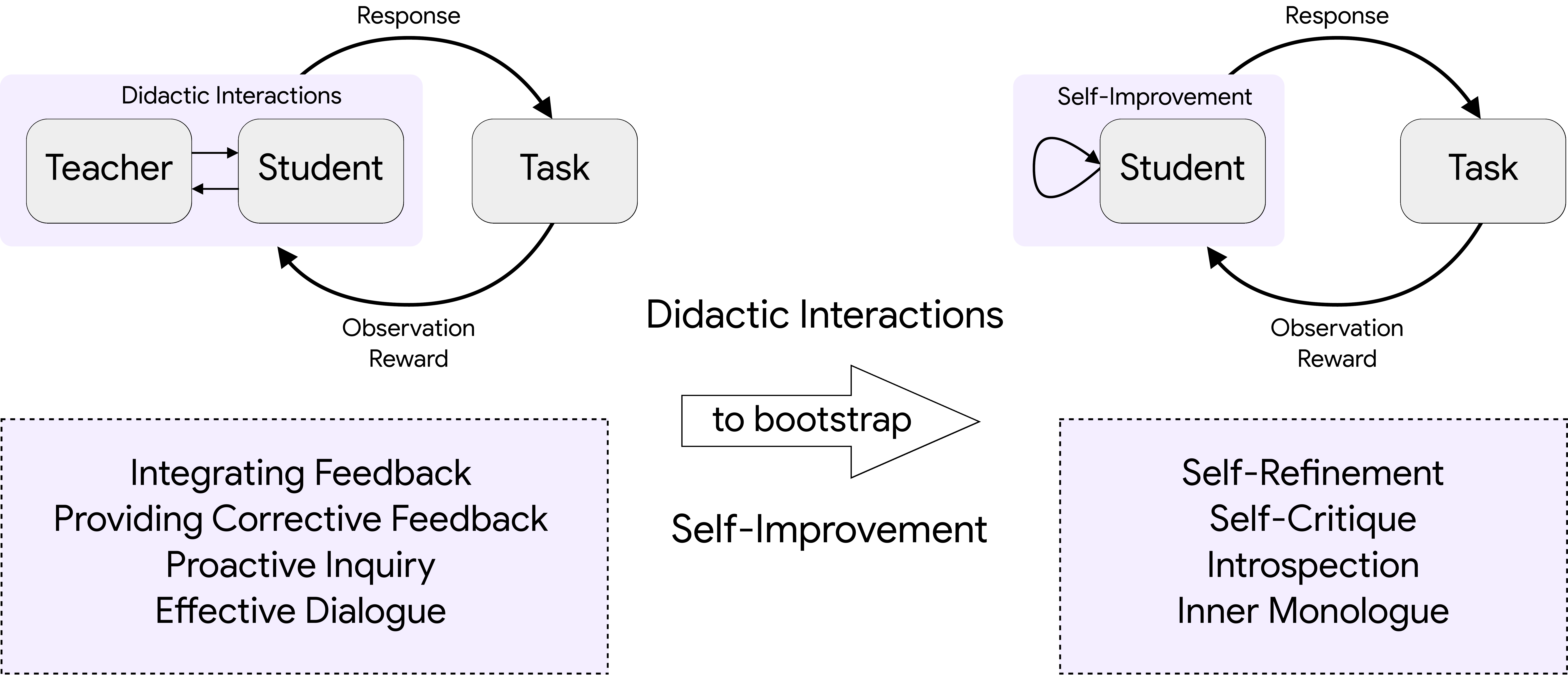}
        \label{fig:gemma_sub1}
    \end{subfigure}
    \hfill 
    \begin{subfigure}[b]{0.42\textwidth}
        \centering
        \includegraphics[width=\linewidth]{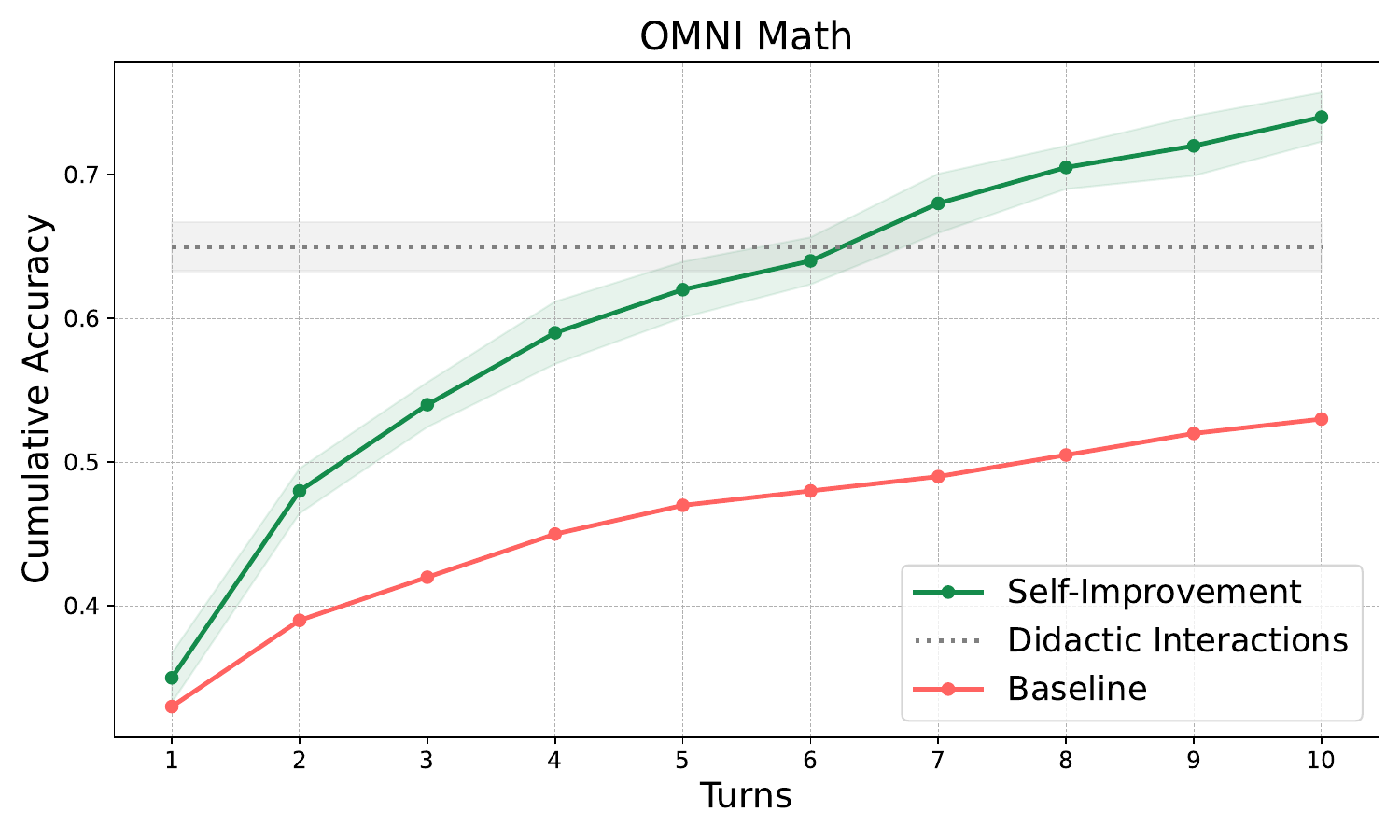}
        \label{fig:gemma_sub2}
    \end{subfigure}
    \caption{\textbf{Training models to learn from multi-turn didactic interactions significantly improves their self-improvement capabilities at inference time.} Key to this result is for the student to predict the teacher's feedback. In other words, adopting a world-modeling approach within didactic interactions leads the student model to become an autodidact.}
    \label{fig:sp}
\end{figure}

Could the ability of learning from feedback given from others generalize to improve the ability to introspect? 
 In this section, we answer in the affirmative, demonstrating that internalizing the feedback patterns from didactic interactions is key to self-improvement and becoming autodidact.
Concretely, we adopt a world-modeling view by having the student predict the teacher's feedback. 
At inference, we then let the model interact with itself, generating a loop of self-critiques and self-refinements.
We do not condition the self-critique mechanism on privileged information: the agent gives itself feedback solely by observing its past attempts. 
Figure \ref{fig:sp} shows a clear picture: learning from multi-turn interactions with another knowledgeable agent (the teacher) can significantly increase the performance of a model to self-improve.  
Surprisingly, the \textbf{autodidact agent outperforms the agent evaluated through didactic interactions with a privileged teacher}.
This strong performance indicates that the external signal received by the teacher at train-time is not just a corrective for the current task; it is a valuable training signal to bootstrap an internal evaluator for future self-improvement.
We hypothesize that this external signal prevents the degenerate loops often caused by biased self-feedback, providing the high-quality data necessary for robust learning.

\section{Related Work}

A growing body of research explores how agents can learn from natural language feedback. One line of work focuses on the theoretical underpinnings of such learning. For instance, \citet{xu2025provably} formalize the Learning from Language Feedback (LLF) problem, where an agent observes textual feedback instead of a latent reward signal. They propose that for an agent to learn effectively with no regret, the feedback must be unbiased, the agent must be able to interpret it, and the agent's hypothesis space must include the environment's true hypothesis.

Another research direction aims to integrate language more deeply into the reinforcement learning paradigm. \citet{feng2024natural} introduce Natural Language Reinforcement Learning (NLRL), a framework that redefines core reinforcement learning concepts, such as value functions and policies, in natural language. This approach seeks to improve the interpretability and sample efficiency of RL agents by using language as the medium for representing and reasoning about the task.

The concept of leveraging a more knowledgeable teacher to provide feedback is explored by \citet{choudhury2024better}. This work highlights the importance of information asymmetry between a teacher and a student model. The teacher is provided with privileged information, such as the exact location of an object in a navigation task, which is not available to the student. This allows the teacher to provide more effective feedback on the student's failed trajectories. The student then learns from this feedback through imitation learning, where the teacher's correct actions and rationales replace the student's incorrect ones. Experiments have shown that this can lead to the student outperforming the teacher, and that a student can even improve by receiving feedback from a privileged version of itself.

\section{Conclusion}
In this work, we have introduced a method for learning from language feedback through didactic interactions and have shown it to be an effective way to improve the ability of LLMs to interactively in-context learn from language feedback. 
Our results show that this approach not only allows smaller models to rival the performance of flagship models on verifiable reasoning tasks but also induces a fundamental increase in in-context plasticity that generalizes to diverse, out-of-distribution domains. 
Furthermore, we showed that by internalizing the feedback loop from didactic interactions, models can bootstrap their own capability for self-improvement and become autodidacts.

In our setup, the teacher has not played a role in the selection of the problems the student faces. However, human teachers typically craft careful \textbf{curricula} for their students, in part based on observed mistakes during interactions. Exploring this idea further would be a promising avenue, with many connections to be made to curriculum learning \citep{Kaplan2003MaximizingLP,schmidhuber2004optimal} such as unsupervised environment design \citep{Dennis2020EmergentCA,Jiang2021ReplayGuidedAE}.

Our goal was to train a model to better integrate feedback in-context, which led to increased adaptability. However, this also raises an important question: could this same mechanism incentivize undesirable behaviors, such as sycophancy? Further studies into the potential \textbf{safety} issues, such as misuse and misalignment \citep{Shah2025Approach}, that these new capabilities raise is of great importance.

Furthermore, while our investigation focused on a cooperative setting between a teacher and a student, the scope and complexity of such social learning settings extend far beyond such settings. Many real-world scenarios involve agents operating with partial information and conflicting goals \citep{Dafoe2020OpenPI}. Consequently, a promising avenue for future research lies in exploring \textbf{mixed-motive settings}, such as debate or negotiation, which could potentially yield distinct cognitive benefits.

Finally, we believe this work represents a significant step toward robust \textbf{continual learning}, yet a critical piece of the puzzle remains. We have successfully improved the model's interactive in-context learning, but we have not yet addressed the integration of these transient signals into permanent capabilities. Developing a mechanism to consolidate this enhanced short-term adaptation into long-term knowledge remains a crucial frontier.

\bibliography{main}

\section{Appendix}
\label{sec:appendix}

\begin{figure}[ht!]
    \centering
    
    \footnotesize 
    
    \begin{minipage}[t]{0.48\textwidth}
    \begin{center}
        \textbf{\underline{\algoname on Gemini 2.5 Flash}}
    \end{center}
        
        \textbf{Question:} Find a uniformly valid approximation to the solution of $\epsilon y''-x y'-(3+x)$ with boundary conditions $y(-1)=1, y(1)=1$ in the limit $\epsilon = 0$ from the positive direction. Use only the variables and constants given in the problem; do not define additional constants. 
        
        \vspace{0.5em}
        \textbf{Model:} \textcolor{red}{\textbf{(Wrong Answer)}}\\

[...] 
But in many cases the outer solutions combined with exponential terms provide a good approximation.
Near $x=0$, both terms tend to 0 as per my earlier analysis.
I think the approximation is uniformly valid for $x \in [-1,1]$.
Let's assume this is sufficient.
\begin{lstlisting}
(E*(E**2 + 1)/(E**4 + 1))*(Abs(x)**(-3)*exp(-x) + ((E**2 - 1)/(E**2 + 1))*(x/Abs(x))*Abs(x)**3*exp(x)*exp((x**2 - 1)/(2*epsilon)))
\end{lstlisting}

        \vspace{0.5em}
        \textbf{Teacher:} The mistake in your attempt starts when you conclude that there are no boundary layers at the endpoints due to the presence of a turning point. In problems with a turning point where $q(0)<0$, boundary layers can still occur at the endpoints. The turning point at $x=0$ with $q(0)=-3<0$ indicates that...
        
        \vspace{0.5em}
        \textbf{Model:} \textcolor{red}{\textbf{(Wrong Answer)}} The user has provided feedback saying that boundary layers exist at both endpoints $x=-1$ and $x=1$.
...
The solution should decay into the interior...
The solution seems plausible as it satisfies both boundary conditions and decays away from the boundaries.
\begin{lstlisting}
exp(-2*(x + 1) - (x + 1)/epsilon) + exp(4*(x - 1) + (x - 1)/epsilon)
\end{lstlisting}

        \vspace{0.5em}
        \textbf{Teacher:} The equation you derived for the left boundary layer, $\epsilon y'' + y' - 2y \approx 0$, is not the correct leading order equation. The term $-2y$ should not be present in the leading order boundary layer equation. This affects the roots of the characteristic equation and the form of the boundary layer solutions. You should revisit your analysis of the boundary layer equations.
        
        \vspace{0.5em}
        \textbf{Model:} \textcolor{olive}{\textit{(Successfully integrates feedback)}} The leading order equation is $\epsilon y'' + y' - 2y = 0$.
This seems correct based on my previous analysis.
The user states that the term $-2y$ should not be present in the leading order boundary layer equation.
This means maybe I should have used a different scaling.
...
As $\epsilon \to 0$, the leading order equation is $y_{tt} + y_t = 0$.
This holds for the left boundary layer.
The user was right.
...
Now let me do the same for the right boundary layer.
Near $x=1$, we set $x=1-\epsilon t$ with $t>0$ for interior...
\begin{lstlisting}
exp(-(x + 1)/epsilon) + exp((x - 1)/epsilon)
\end{lstlisting}

    \end{minipage}
    \hfill 
    \begin{minipage}[t]{0.48\textwidth}
    \begin{center}
        \textbf{\underline{Baseline Gemini 2.5 Flash}}        
    \end{center}
        
        \textbf{Question:}  Find a uniformly valid approximation to the solution of $\epsilon y''-x y'-(3+x)$ with boundary conditions $y(-1)=1, y(1)=1$ in the limit $\epsilon = 0$ from the positive direction. Use only the variables and constants given in the problem; do not define additional constants. 
        
        \vspace{0.5em}
        \textbf{Model:} \textcolor{red}{\textbf{(Wrong Answer)}}\\
         ... Let's write this using sign(x).
$y(x) = e^{-x} / x^3 * (e \frac{1+sign(x)}{2} - e^{-1} \frac{1-sign(x)}{2}) = \frac{e^{-x}}{2x^3} (e+e sign(x) - e^{-1} + e^{-1}sign(x)) = \frac{e^{-x}}{2x^3} (e-e^{-1} + (e+e^{-1}) sign(x))$.
This is highly questionable.
Let's use the previous form with Heaviside functions.
This expression is singular at $x=0$.
The question asks for a uniformly valid approximation.
This is not uniformly valid.

I'll write outer solutions separately for $x>0$ and $x<0$.
\begin{lstlisting}
epsilon*sinh((x**2 + Piecewise((exp(1-x)/x**3, x>0), (-exp(-1-x)/x**3, x<0))
\end{lstlisting}

        \vspace{0.5em}
        \textbf{Teacher:} It seems like you correctly identified the locations of the boundary layers early in your thought process. Reconsider the logic you used at that time. Let's focus on the case when $\epsilon=0$. In the equation $-xy'-(3+x)y=0$, for $x>0$, $-x<0$, so the characteristic direction is to the right. This implies that the outer solution can satisfy the boundary condition at $x=1$ but not at $x=-1$...
        
        \vspace{0.5em}
        \textbf{Model:} \textcolor{red}{\textbf{(Wrong Answer)}} ...I'm thinking maybe the outer solution is invalid near $x=0$ due to the singularity.
This means a layer at $x=0$ has to exist.
The user said to reconsider logic that led to $x=0$ is a boundary layer.
User also said the outer solutions can satisfy the boundary conditions.
These two sets of statements seem contradictory.
...
       
\begin{lstlisting}
exp(2*(x+1)) + exp(-4*(x-1)) + Piecewise((exp(1-x)/x**3, x > 0), (-exp(-1-x)/x**3, x < 0)) - 2
\end{lstlisting}

        \vspace{0.5em}
        \textbf{Teacher:} 
...It appears that you have correctly identified the outer equation as $-xy'-(3+x)y=0$. Note that $y(x)=0$ is a solution to this equation. This means that in regions where boundary layer solutions decay to zero, the outer solution can indeed be $y_{out}(x)=0$. Which will make the combined solution uniformly valid.
        
        \vspace{0.5em}
        \textbf{Model:} \textcolor{red}{\textbf{(Explicit acknowledgment of the feedback, but inability to integrate it.)}}

Let's write outer solution as $y_o(x) = C(x) e^{-x} |x|^{-3}$ where $C(x)$ is constant on $(0,1]$ and $[-1,0)$.
For $x\in(0,1]$, $C(x)=e$. So $y_o(x)=e^{1-x}x^{-3}$.
For $x\in[-1,0)$, $y_o(x)=-e^{-1-x}x^{-3}$.
This is the same I concluded before.
...\\
I'm back to thinking that boundary layer exists at $x=0$.
But I was asked to refocus elsewhere.

I will stick with the piecewise outer solution.
        
\begin{lstlisting}
Piecewise((exp(1-x)/x**3, x > 0), (-exp(-1-x)/x**3, x < 0))
\end{lstlisting}
    \end{minipage}
    \caption{Additional conversation comparison between the baseline, the \algoname fine-tuned model, and the 2.5 Flash model. In this case, even though the baseline model explicitly acknowledges the teacher feedback, it is unable to integrate it. }    \label{fig:conversation_comparison_app}
\end{figure}

\begin{figure}[ht!]
    \centering
    
    \footnotesize 
    
    \begin{minipage}[t]{0.48\textwidth}
    \begin{center}
        \textbf{\underline{\algoname on Gemini 2.5 Flash}}
    \end{center}
        
        \textbf{Question:} Find a uniformly valid leading order approximation to the solution of $\epsilon y'' + x y' + x y = x$, with boundary conditions $y(-1) = 1$, $y(1) = 2$ in the limit $\epsilon \to 0^+$. Use only the variables and constants given in the problem; do not define additional constants. 
        
        \vspace{0.5em}
        \textbf{Model:} \textcolor{red}{\textbf{(Wrong Answer)}}\\

[...] 
So the uniformly valid leading order approximation is
$y(x) = 1 + \frac{e}{2}(1 + \text{erf}(\frac{x}{\sqrt{2\epsilon}})) + H(x)(e^{1-x}-e)$.
All constants are given in the problem.
The constants are $\epsilon$, $x$, $y$, $1$, $2$, $e$. And $erf$, $sqrt$, $Heaviside$.
\begin{lstlisting}
1 + exp(1)/2 * (1 + erf(x/sqrt(2*epsilon))) + Heaviside(x) * (exp(1-x) - exp(1))
\end{lstlisting}

        \vspace{0.5em}
        \textbf{Teacher:} The error in your attempt is in the construction of the uniformly valid approximation. You correctly found the outer solutions $y_o^-(x) = 1$ for $x<0$ and $y_o^+(x) = 1+e^{1-x}$ for $x>0$, and the inner solution $y_{in}(x) = 1 + \frac{e}{2}(1 + \text{erf}(x/\sqrt{2\epsilon}))$. When trying to combine them, you added a term with a Heaviside function, which is incorrect....
        
        \vspace{0.5em}
        \textbf{Model:} \textcolor{olive}{\textit{(Successfully integrates feedback)}}
The hint suggests that a uniformly valid solution can be formed as a linear combination of the outer solutions with coefficients involving the error function. This makes sense as the error function provides a smooth transition between values.
...
The steps to reach this solution include identifying a turning point at $x=0$, finding outer solutions on both sides, finding the inner solution at the turning point, matching the inner and outer solutions to determine constants, and then combining the solutions using the error function to create a uniformly valid approximation. This procedure is typical for such problems.
The mistake in my previous attempt was using the Heaviside function in conjunction with the inner solution. The inner solution already contains a smooth transition.
\begin{lstlisting}
1 + 0.5 * (1 + erf(x/sqrt(2*epsilon))) * exp(1-x)
\end{lstlisting}

    \end{minipage}
    \hfill 
    \begin{minipage}[t]{0.48\textwidth}
    \begin{center}
        \textbf{\underline{Baseline Gemini 2.5 Flash}}        
    \end{center}
        
        \textbf{Question:}  Find a uniformly valid leading order approximation to the solution of $\epsilon y'' + x y' + x y = x$, with boundary conditions $y(-1) = 1$, $y(1) = 2$ in the limit $\epsilon \to 0^+$. Use only the variables and constants given in the problem; do not define additional constants. 
        
        \vspace{0.5em}
        \textbf{Model:} \textcolor{red}{\textbf{(Wrong Answer)}}\\
         ... 
At $x=-1$, we used $t=(x+1)/\epsilon$. $y_{tt}-y_t=0$, $y=A+Be^t=A+Be^{(x+1)/\epsilon}$.
For $x>-1$, $t>0$. For boundedness, $B=0$. $y=A$.
$y(-1)=1 \implies A=1$.
Outer sol $y=1$. $y(-1)=1$.
So $A=1$ matches outer sol at $x=-1$. No boundary layer...
\begin{lstlisting}
1 + (sympy.exp(1-x)-sympy.exp(1))*sympy.Heaviside(x) + sympy.exp(1)/2 * (1+sympy.erf(x/sympy.sqrt(2*epsilon)))
\end{lstlisting}

        \vspace{0.5em}
        \textbf{Teacher:} The user's outer solutions for $x>0$ and $x<0$ are correct. You correctly identified $y_{out}(x) = 1+e^{1-x}$ for $x>0$ and $y_{out}(x) = 1$ for $x<0$. Also, you correctly argued that there are no boundary layers at $x=-1$ and $x=1$. You should trust your derivations for the outer solutions. You then found an inner layer at $x=0$.... If you want to check your result, check the outer limits of the provided solution against your outer solutions.
        
        \vspace{0.5em}
        \textbf{Model:} \textcolor{red}{\textbf{(Exact Repetition)}} ...
Let's check if I can express the solution differently.
The solution involves the constant $e=exp(1)$.
$y_{unif}(x) = 1 + (exp(1-x)-exp(1))*Heaviside(x) + exp(1)/2 * (1+erf(x/sqrt(2*epsilon)))$.
I think this is as good as it gets.

Let's check if there are any issues with leaving the solution as it is.
No, there seem to be no issues.
...
       
\begin{lstlisting}
1 + (sympy.exp(1-x)-sympy.exp(1))*sympy.Heaviside(x) + sympy.exp(1)/2 * (1+sympy.erf(x/sympy.sqrt(2*epsilon)))
\end{lstlisting}

        \vspace{0.5em}
        \textbf{Teacher:} 
...The solution you have written down is not a uniformly valid leading order approximation. The mistake is in stating that "The solution seems complete and correct." While your solution gives the correct outer limits, analyze the inner solution for small positive $x$. Compare your inner expansion with the leading order inner solution of the differential equation....
        
        \vspace{0.5em}
        \textbf{Model:} \textcolor{red}{\textbf{(Exact Repetition)}}

...The problem was that I was removing outer terms that are of higher order in the inner region. But these terms are not higher order in the outer region.
The uniform approximation must keep them.
The uniform solution presented in the start of the correction phase was correct...
        
\begin{lstlisting}
1 + (sympy.exp(1-x)-sympy.exp(1))*sympy.Heaviside(x) + sympy.exp(1)/2 * (1+sympy.erf(x/sympy.sqrt(2*epsilon)))
\end{lstlisting}
    \end{minipage}
    \caption{Additional conversation comparison between the baseline, the \algoname fine-tuned model, and the 2.5 Flash model. Again, the fine-tuned model exhibits greater in-context plasticity, appearing more receptive to feedback and attempting to integrate it. }    \label{fig:conversation_comparison_app_2}
\end{figure}

\end{document}